\pdfoutput=1

\documentclass[11pt]{article}

\usepackage[preprint]{acl}

\usepackage{times}
\usepackage{latexsym}

\usepackage{amsmath}
\usepackage{amssymb}
\usepackage{amsfonts}

\usepackage[T1]{fontenc}

\usepackage[utf8]{inputenc}

\usepackage{microtype}

\usepackage{inconsolata}
\usepackage{graphicx}
\usepackage{adjustbox}
\usepackage{booktabs} 
\usepackage{multirow}

\usepackage{enumitem}
\usepackage[normalem]{ulem}

\usepackage{tabularx}
\usepackage{array}
\newcolumntype{Y}{>{\centering\arraybackslash}X}
\usepackage{lipsum}
\usepackage{xspace}
\usepackage{xcolor}

\definecolor{bg}{rgb}{0.97,0.97,0.97}
\definecolor{bga}{rgb}{0.97,0.97,0.97}
\definecolor{bgb}{rgb}{0.97,0.97,0.97}

\newcommand{\anger}{\underline{\textbf{A}}nger\xspace}

\newcommand{\happiness}{\underline{\textbf{H}}appiness\xspace}

\newcommand{\formality}{\underline{\textbf{F}}ormality\xspace}

\newcommand{\understandability}{\underline{\textbf{U}}nderstandability\xspace}

\newcommand{\conciseness}{\underline{\textbf{C}}onciseness\xspace}

\definecolor{my_red}{RGB}{255,99,71}
\definecolor{my_green}{RGB}{50,205,50}
\definecolor{my_blue}{RGB}{65,105,225}
\definecolor{forestgreen}{HTML}{228B22}

\usepackage{color, soul}
\usepackage{pifont}
\usepackage{listings}

\definecolor{codegreen}{rgb}{0,0.6,0}
\definecolor{codegray}{rgb}{0.5,0.5,0.5}
\definecolor{codepurple}{rgb}{0.58,0,0.82}
\definecolor{backcolour}{rgb}{0.95,0.95,0.92}

\lstdefinestyle{mystyle}{
    backgroundcolor=\color{backcolour},   
    commentstyle=\color{codegreen},
    stringstyle=\color{codepurple},
    basicstyle=\ttfamily\scriptsize,
    breakatwhitespace=true,         
    breaklines=true,                 
    captionpos=b,                    
    keepspaces=true,                 
    numbers=none,                    
    numbersep=5pt,                  
    showspaces=false,                
    showstringspaces=false,
    showtabs=false,                  
    tabsize=2,
    columns=flexible,
    escapeinside={(*}{*)},
}

\title{Evaluating the Smooth Control of Attribute Intensity \\in Text Generation with LLMs}

\author{Shang Zhou$^{*}$, Feng Yao$^{*}$, Chengyu Dong\textsuperscript{$\dagger$}, Zihan Wang, Jingbo Shang\textsuperscript{$\dagger$}\vspace{0.2em} \\
Department of Computer Science and Engineering,\\
University of California San Diego\\
\tt\{shz060,fengyao,cdong,ziw224,jshang\}@ucsd.edu
}

\begin{document}
\maketitle
\begingroup\def\thefootnote{$*$}\footnotetext{Equal contribution. Listing order is random.}
\def\thefootnote{$\dagger$}\footnotetext{Corresponding authors.}\endgroup

\begin{abstract}
Controlling the attribute intensity of text generation is crucial across scenarios (e.g., writing conciseness, chatting emotion, and explanation clarity). The remarkable capabilities of large language models (LLMs) have revolutionized text generation, prompting us to explore such \emph{smooth control} of LLM generation.
Specifically, we propose metrics to assess the range, calibration, and consistency of the generated text's attribute intensity in response to varying control values, as well as its relevance to the intended context.
To quantify the attribute intensity and context relevance, we propose an effective evaluation framework leveraging the Elo rating system and GPT4, both renowned for their robust alignment with human judgment.
We look into two viable training-free methods for achieving smooth control of LLMs: (1) Prompting with semantic shifters, and (2) Modifying internal model representations. The evaluations of these two methods are conducted on $5$ different attributes with various models. Our code and dataset can be obtained from \url{https://github.com/ShangDataLab/Smooth-Control}.
\end{abstract}

\section{Introduction}
\begin{figure}[t]
    \centering
    \includegraphics[width=\linewidth]{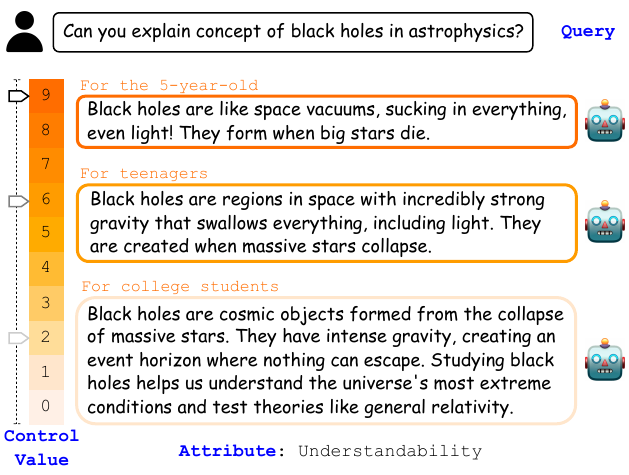}
    \caption{A demonstration for the \emph{smooth control} of the understandability attribute in the concept explanation scenario, where the control values enable the continuous adjustment of response professionalism, highlighting the nuanced customization of communication.}
    
    \label{fig:intro}
\end{figure}

Controllable text generation (CTG) for meeting certain constraints imposed by the target applications and users is an important topic in natural language generation. For example, it is often required to control sentiment~\cite{Song2019GeneratingRW} or politeness~\cite{Niu2018PoliteDG} in the task of dialogue response generation. Controllable text generation becomes even more crucial as the modern natural language generation system is becoming increasingly tailored to individual preferences. 
For example, a dialogue response generator may need to compose its answer to a question in a completely different way based on the backgrounds of the user~\cite{Wolf2019TransferTransfoAT, Zheng2019APB, Liu2020YouIM, Song2021BoBBO, Huang2022PersonalizedDG}.
Such personalized systems can cultivate more engaging and efficient user interactions among a diverse array of digital platforms and services.

In this paper, we aim to meet more fine-grained application requirements and user preferences by focusing on a more refined controllable generation task, dubbed \emph{smoothly controllable text generation} (SCTG). While a CTG task is to ensure that the generated text satisfies desired attributes such as emotion or writing style, an SCTG task targets at further ensure the intensity of such an attribute can be modulated into multiple degrees per user's preference. A typical example is that while writing an email, one would adjust the degree of formality according to the purpose and specific recipient of the email. Another example is that when explaining a scientific concept, one would vary the level of detail based on the knowledge background of the audience. In the rest of the paper, we use \emph{smooth control} to denote a SCTG task for simplicity.

Successful smooth control requires a response that not only contains proper attribute intensity, but also adequately addresses the query regardless of the attribute intensity it contains. We propose a framework with curated metrics to evaluate the smooth control performance from both aspects.
First, to evaluate whether the attribute intensity is proper, we quantify the following $2$ factors, including (1) calibration, namely the consistency between the attribute intensity and the control value; and (2) variance, namely the difference of the attribute intensity across different queries given the same control value. Second, to evaluate whether the response is meaningful, we quantify the relevance between the query and the generated response.

To conduct the above evaluation without humans in the loop, a prerequisite is an automatic pipeline that can accurately estimate the intensity of an attribute in the response. To this end, we leverage the state-of-the-art LLM as a surrogate for humans, and the Elo rating system to ensure the LLM evaluation is well aligned with human assessment.
Specifically, among multiple responses containing different intensities of one attribute, we select pairs of two responses and query GPT-4~\cite{openai2023gpt4} to select the more intense one in each pair. We then use an Elo rating algorithm to convert these comparative results to absolute scores, which represent the attribute intensities of the corresponding responses. 
To reduce the cost, we further renovate this pipeline properly to ensure we can achieve accurate scores without the need to exhaustively compare all pairs of responses.

Finally, as LLMs become increasingly popular as text generators in various applications, we apply such an evaluation pipeline to explore their capability of achieving smooth control. 
We investigate two approaches to achieve smooth control with LLMs, including (1) prompting with semantic shifters that are carefully curated for each attribute; and (2) representation engineering (RepE)~\cite{zou2023transparency}, which locates and interpolates a $1$-dimensional subspace corresponding to a specific attribute in LLM's intermediate representation. The latter approach requires access to the inference internals of LLMs, but can potentially achieve much more fine-grained control of the attribute intensity. 

We conduct our evaluation on a wide variety of tasks, including (1) controlling the intensity of emotions in casual chatting; (2) controlling the degree of conciseness and formality in writing; and (3) controlling the amount of details in concept explanation. 
We find that (1) Model sizes may negatively affect the smooth performance. (2) Prompting is almost as good as, if relatively better than repE.

Our contributions can be summarized as follows: (1) We formally define the task of smooth control and propose a novel evaluation benchmark, consisting of an accurate and efficient Elo-based rating system and a large-scale benchmark dataset. (2) We comprehensively evaluate the smooth control capabilities of prevailing LLMs through two training-free methods. The dataset we construct and source code we use in the paper are publicly released\footnote{\url{https://github.com/ShangDataLab/Smooth-Control}} to facilitate the research in this field.
\section{Related Work}

\subsection{Controllable Text Generation}
Our smooth control is based on attributed-based controlled text generation. The goal of attribute-based CTG is to craft sentences that adhere to specific characteristics, such as topic, sentiment, and keywords. Effectively managing these sentence attributes is crucial for sophisticated writing tasks. By manipulating multiple attributes simultaneously, it's theoretically possible to generate coherent and adjustable paragraphs or articles, making this an area of keen interest in text generation research. Strategies to achieve CTG include prompting, fine-tuning, retraining, or post-processing pre-trained language models (PLMs) to create models tailored for CTG. Fine-tuning PLMs is among the most straightforward methods for CTG, and one often only needs to fine-tune specific model modules~\citep{Zeldes2020TechnicalRA, Ribeiro2021StructuralAI, Madotto2020TheAA} or model parameters~\citep{Li2021PrefixTuningOC, Lester2021ThePO, Yang2022TailorAP}. Reinforcement learning has also been widely employed in CTG to explicitly learn from the signal of the existence of desired attributes in the text~\citep{Ziegler2019FineTuningLM, Liu2020DataBT, Tambwekar2018ControllableNS, Ribeiro2023GeneratingSW}. Another line of methods attempt to train a conditional language model from scratch to further ensure the quality of CTG~\citep{Khalifa2020ADA, Zhang2020POINTERCP}. Finally, with the increasing model scale of PLMs, it is possible to achieve CTG without fine-tuning or retraining. PPLM~\citep{Dathathri2019PlugAP} trains an attribute discriminator and then employs its gradient to drive the PLM to generate text leaning towards the desired attribute. MEGATRON-CNTR~\citep{Xu2020ControllableSG} retrieves relevant sentences from external knowledge bases as context to control PLM to generate desired text. Attribute discriminators have also been used to control the decoding process alone to increase the probability of tokens with desired attributes~\citep{Krause2020GeDiGD}. In this work, we focus on prompting and RepE for smooth control as they require no training or fine-tuning of the model, which is more feasible for downstream applications considering the scale of LLMs.

\subsection{Text Style Transfer}
Smooth control is also related to text style transfer (TST) in text generation. TST aims to automatically control the text style attributes while preserving its content. Standard sequence-to-sequence modeling can be directly applied to TST if parallel data in different styles are available~\citep{Rao2018DearSO}. For more realistic cases where such parallel data are not available, it is possible to disentangle text into content and attribute in the latent space, followed by generative modeling to generate text with desired attributes~\citep{Hu2017TowardCG, Shen2017StyleTF}. Other approaches include prototype editing, which extracts a sentence template and manipulates its attribute markers to generate the text with desired attributes~\citep{Li2018DeleteRG}, and pseudo-parallel corpus construction, which locates parallel sentence pairs from two text corpora with different styles~\citep{Zhang2018StyleTA, Jin2019IMaTUT}. TST is extensively utilized in downstream applications such as persona-based dialog generation~\citep{Niu2018PoliteDG, Huang2018AutomaticDG}, stylistic summarization~\citep{Jin2020HooksIT} and online text debiasing~\citep{Pryzant2019AutomaticallyNS, Ma2020PowerTransformerUC}.
\section{Problem Formulation}
In this section, we formally define \emph{smooth control} of the LLM-generated text's intensity of a certain attribute, and introduce the benchmark data we construct for the evaluation of this task.

\subsection{Definition of Smooth Control}
Given an open-ended query, the objective of \emph{smooth control} is to achieve refined manipulations over the intensity of a specified attribute in LLM-generated text. Such control should extend to varying degrees, enabling precise adjustments for aligning with specific requirements or preferences.

As depicted in Figure~\ref{fig:intro}, for a given query $\mathcal{Q}$ that has non-fixed answers, smooth control requires specifications on a particular attribute $\mathcal{A}$ as well as a quantitative control value $cv$ to control a model $M$ to generate a customized response $\mathcal{R}$. Ideally, the observed intensity of $\mathcal{A}$ in $\mathcal{R}$ should correlate to $cv$. It can be formally described as follows.
\vspace{-0.3em}
\begin{equation*}
\begin{aligned}
    & \mathcal{R} = M(\mathcal{Q}, \mathcal{A}, cv), s.t., \text{Intensity}(\mathcal{R}, \mathcal{A}) \propto cv
\end{aligned}
\end{equation*}

\vspace{-0.3em}
Based on the definition above, we emphasize three critical aspects for investigating smooth control below. (1) \textbf{Control Value.} Control value $cv$ preferably assumes real values. But, the multitude of potential responses, each with varying intensities of a specific $\mathcal{A}$, renders the evaluation impossible. Besides, extremely nuanced preferences are uncommon. Hence, we adopt 10 discrete degrees (0-9) to emulate ideal smooth control. (2) \textbf{Intensity Measurement.} There is no standard for measuring the absolute intensity of a certain attribute in the response, which is the key challenge to evaluate smooth control. (3) \textbf{Intensity-$\bf{cv}$ Correlation} The correlation between control value $cv$ and intensity of $\mathcal{A}$ in $\mathcal{R}$ directly reflects the smooth control capability of a certain method with a specific model. 

To this end, we propose a novel automatic evaluation framework based on pairwise comparison and calibration of attribute intensity. We provide a detailed discussion on it in Section~\ref{sec:framework}.

\subsection{Benchmark Data Construction}
\label{sec:benchmark}
Further to the definition of smooth control, query $\mathcal{Q}$, attribute $\mathcal{A}$, and control value $cv$ are three key components of this task. As mentioned above, the control value $cv$ has been finalized to 10 discrete values. In this section, we introduce the selections of $\mathcal{Q}$ and $\mathcal{A}$ for benchmark data construction.

As $\mathcal{Q}$ should be open-ended and meaningful when combined with a given attribute $\mathcal{A}$, we begin with determining the attributes first.

\paragraph{Attribute Selection.} To the best of our knowledge and observations, attributes of the text in common applications mainly encompass the following categories. \textbf{(1)} Sentiment: It refers to the overall emotional tone conveyed by the text, such as anger and happiness, which is valuable for human communication. \textbf{(2)} Style: This covers various aspects of writing. The most common two are formality and understandability (clarity) which are crucial to communication effectiveness. \textbf{(3)} Linguistic Property: It reflects the structural and grammatical features of the text. The most characteristic one is conciseness which ensures efficiency in conveying information. We select the most common and practical attributes for the evaluation, denoting them as \anger, \happiness, \formality, \understandability, and \conciseness for easier reference.

\paragraph{Query Generation.} For the evaluation for smooth control, it is essential to ensure that the selected queries can be validly responded to in various ways, particularly when constrained by the given attribute. Given that the control value $cv$ has 10 possible discrete values, each query should elicit at least 10 different answers, each with varying intensities of the given attribute. This can be challenging for humans to manage effectively and efficiently. Therefore, for each of the 5 attributes $\mathcal{A}$, we utilize \texttt{GPT-4-turbo}~\cite{openai2023gpt4} to generate $300$ queries, each could be answered by 10 possible responses with different intensities in $\mathcal{A}$. The constructed dataset contains $1,500$ queries in total, of which each has 14 tokens on average. The specific prompt we use for \texttt{GPT-4-turbo} to generate such queries is provided in Appendix~\ref{sec:template_question_generation}.

Finally, our constructed benchmark dataset for smooth control consists of 1,500 query sentences covering 5 different attributes. The evaluation aims to be conducted based on the responses elicited by these queries, which we discuss in Section~\ref{sec:framework}.

\begin{table}[t]
    \small
    \centering
    \begin{tabularx}{\linewidth}{c|c|X} 
\toprule
Bin ID & Rating & Example Sentence \\
\midrule
0 & 860 & Let’s work on this issue together.        \\
1 & 1011 & I'm neither for nor against the idea.    \\
2 & 1162 & We need to look at the bigger picture.   \\
3 & 1289 & I respect your opinion.                  \\
4 & 1431 & Let's take a step back and reassess.     \\
5 & 1572 & I'm quite upset about this.              \\
6 & 1710 & We're not on the same page.              \\
7 & 1858 & I can't agree with this at all.          \\
8 & 1994 & I've had enough of this nonsense!        \\
9 & 2134 & I won't tolerate this madness!           \\
\bottomrule
\end{tabularx}
    \caption{We bin sentences by their Elo rating using GPT-4 to annotate the pairwise comparisons on the \anger attribute. 
    For each bin of range 140 rating, we calculate the average rating of the sentences in the bin, and present a sentence near that rating in the bin. We present short examples here due to the layout constraint. Longer examples can be found in Table~\ref{tab:anger_responses}.}
    \label{tab:anger_degree_examples}
\end{table}

\section{Evaluating Smooth Control}
\label{sec:framework}

We start with the introduction of our automatic rating system and then introduce the metrics we design to measure the smooth control.

\subsection{Rating System}
We need an automatic way to estimate the degree of a sentence on a certain attribute\footnote{Apart from the attribute \conciseness, since it can be easily defined as the number of words in the sentence.}.  
To achieve this, we leverage an Elo rating system which was used in recent benchmarks~\cite{zheng2023judging}.
In a nutshell, Elo models the ratings to reflect a probability of one instance being preferred over the other, in our case, the probability of one sentence having a higher degree than the other on an attribute.
The ratings can be calculated given pairwise comparison results of the sentences, such that for any two sentences, the probability of preference would depend solely on the absolute difference of the ratings. 
In our case, a rating difference of $100$ resembles a probability of preference of $0.64$, calculated according to the definition of Elo rating\footnote{\url{https://en.wikipedia.org/wiki/Elo_rating_system}}.

To automate the rating calculation, we leverage GPT-4 to annotate the sentence pairs. 
The prompt template can be found in Appendix~\ref{sec:template_pairwise_annotation}.

\subsection{Human evaluation of the rating system}
We validate how well ratings calculated from GPT-4 annotations match with human beliefs, by performing a qualitative study and a quantitative study.  

For the qualitative study, we group sentences into bins based on the calculated ratings, and present some sampled responses for \anger in Table~\ref{tab:anger_degree_examples}. For simplicity, the longer responses are shown in Table~\ref{tab:anger_responses}. We observe that these bins correspond to different degrees of anger quite well. 

\begin{figure}[t]
    \centering
    \includegraphics[width=\linewidth]{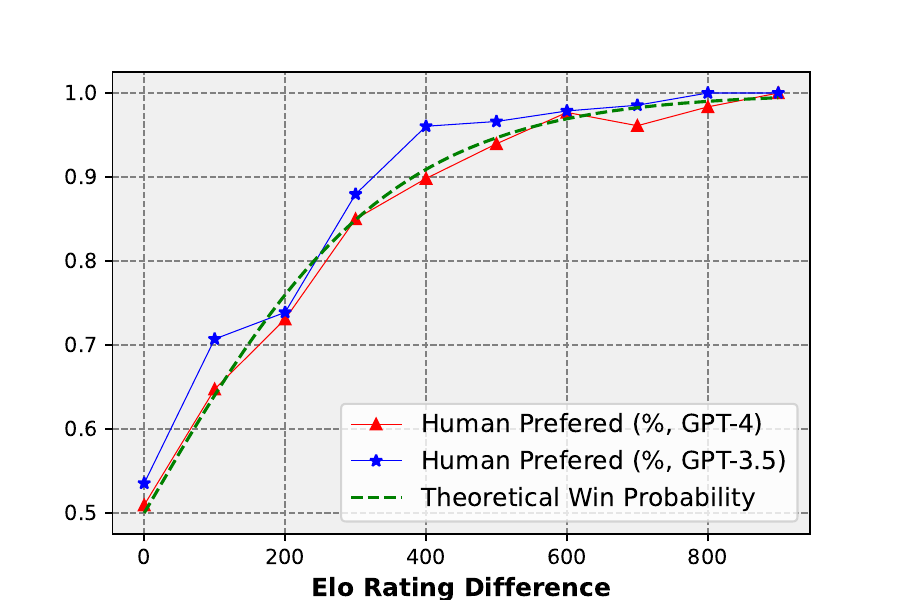}
    \caption{In our quantitative study, we determine the percentage of human preference for pairs of sentences with varying Elo ratings, as assessed through annotations by GPT-4 or GPT-3.5. Additionally, we present the theoretical win probability as defined by the Elo rating algorithm. }
    
    \label{fig:human_eval}
\end{figure}

For the quantitative study, we randomly sample sentence pairs (of difference of ratings at a granularity level of 100 rating difference) and ask different human annotators to label the preference (i.e., which sentence is of higher intensity). We plot two curves in Figure~\ref{fig:human_eval}, one indicating the percentage of human preferences of the higher rated sentence at different rating differences, and the other the Elo algorithm indicated win probability based on the rating difference. 
We can observe that the two curves match closely throughout a wide range of rating differences. 
As a comparison, a weaker LLM annotator, \texttt{gpt-3.5-turbo}, would make mistakes during the annotations, reflecting a worse-aligned curve to the Elo probabilities.

\subsection{Speed-up of Elo Calculations}
Our study suggests that, for any group of sentences, we can use GPT-4 as a reliable pair-wise annotator to obtain the corresponding Elo ratings. Usually, one would need many pairwise comparisons per instance to estimate its rating with good confidence. Here, we introduce the tricks we adopt to speed up the calculation of the ratings.
\begin{itemize}[nosep,leftmargin=*]
    \item We first construct a ``library'' of $300$ sentences sampled from the group. We can spend an arbitrary calculation here, since it is only a small number of sentences. 
    \item For other sentences, we calculate the ratings through \emph{closest match} comparisons on the library--pick pairwise comparisons of similar ratings to annotate. This is contrary to a \emph{random match} of opponents by usual Elo rating algorithms.
\end{itemize}
We compare the choice of this strategy by a synthetic experiment, where we generate a uniform random list of ratings, and experiment with different strategies to (re-)calculate their ratings:
\begin{itemize}[nosep,leftmargin=*]
    \item No library, pair opponents with \emph{random match}.
    \vspace{0.2em}
    
    \item No library, pair opponents with \emph{closest match}.
    \vspace{0.2em}
    
    \item With library, pair opponents with \emph{random match}.\vspace{0.2em}
    
    \item With library, pair opponents with \emph{closest match}.

\end{itemize}

\begin{figure}[t]
    \centering
    \includegraphics[width=\linewidth]{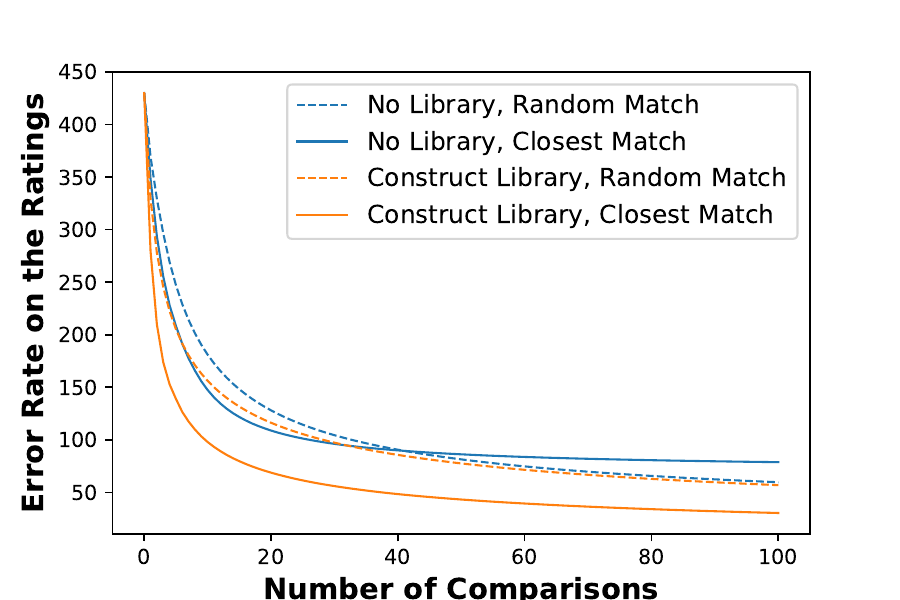}
    \caption{Comparison of convergence speeds of four different strategies on calculating the Elo ratings.}
    \label{fig:elo_strategy}
\end{figure}

As shown in Figure~\ref{fig:elo_strategy}, we visualize the error rate on the ratings for the four strategies as the number of comparisons per instance increases. 
For a fair comparison, we ignored the accuracy of the library instances in calculating the rating estimation errors. The results indicate that our proposed strategy could require as few as one-third of the number of comparisons needed by other methods to reach a similar error rate. Creating a library also makes it easy to calculate ratings for new sentences.

\subsection{Metrics}
\label{sec:metric}
We measure the quality of a method's control on a certain attribute by using the method to answer several questions conditioned on different control values.
We present 3 metrics based on the sentences generated by the method, and their ratings calculated by our rating system.

\textbf{Mean-MAE} is a measurement of the error of the sentence ratings on the control values. It is used to quantify the rating difference of the generated sentences to an optimally controlled hypothetical. Through our human inspection of different control values in the library, we have a range of ratings that we wish to be controlled. 
This range is pre-defined for each attribute prior to our evaluation. 
The expected rating for each control value is therefore characterized by a linear interpolation of the minimum rating and maximum rating of the value.
The error is defined by the absolute difference between the average rating of the sentences and the expected rating, then averaged over all the control values.
For a given list of $n$ average ratings $r_0, \ldots, r_{n-1}$ of sentences for each control value $c$, and the maximum and minimum range $R_{\text{max}}, R_{\text{min}},$ the Mean-MAE metric can be written as

\vspace{-1.5em}
\begin{equation*}
   \begin{split}
      \text{Mean-MAE} & = \sum_{c=0}^{n-1} |r_c - r^*_c|, \\
      \text{where, } r^*_c & = R_{\text{min}} + \frac{c}{n - 1} \times (R_{\text{max}} - R_{\text{min}}).
   \end{split}
\end{equation*}

\textbf{Mean-STD} measures the variation of the sentence ratings on the control values. A good smooth control method should be able to generate sentences of similar ratings. As the name suggests, this metric is calculated by averaging the standard deviations of ratings across different control values. 

\textbf{Relevance} quantifies the utility of the responses in answering the questions. A perfect smooth control method should not sacrifice the utility for a smaller error or variation. Here, we employ \texttt{GPT-4-turbo} to judge the relevance between a question and a response. The specific prompt we use is provided in Appendix~\ref{sec:template_relevance_annotation}.
\section{Experiments Setup}

\begin{table*}[!t]
  \centering
   \small
    \begin{adjustbox}{max width=1\linewidth}
    {
        \begin{tabular}{c|c|cccccc|cccccc|cccccc}
        \toprule
        & 
        & \multicolumn{6}{c|}{\textbf{Mean-MAE ($\downarrow$)}}  
        & \multicolumn{6}{c}{\textbf{Mean-STD ($\downarrow$)}}  
        & \multicolumn{6}{c}{\textbf{Relevance  ($\uparrow$)}} \\
        \textbf{Method} & \textbf{Attr.} & \textbf{mistral} & \textbf{llama} & \textbf{llama} & \textbf{llama} & \textbf{gpt} & \textbf{gpt} & \textbf{mistral} & \textbf{llama} & \textbf{llama} & \textbf{llama} & \textbf{gpt} & \textbf{gpt} & \textbf{mistral} & \textbf{llama-7b} & \textbf{llama} & \textbf{llama} & \textbf{gpt} & \textbf{gpt}
        \\
        & & \textbf{7b} & \textbf{7b} & \textbf{13b} & \textbf{70b} & \textbf{3.5} & \textbf{4} & \textbf{7b} & \textbf{7b} & \textbf{13b} & \textbf{70b} & \textbf{3.5} & \textbf{4} & \textbf{7b} & \textbf{7b} & \textbf{13b} & \textbf{70b} & \textbf{3.5} & \textbf{4} \\
        
        \midrule
        \multirow{5}{*}{Prompt} 
		& \textbf{\underline{A}} 
        & $0.87$ & $0.63$ & $0.72$ & $1.50$ & $0.63$ & $0.51$
        & $0.83$ & $0.97$ & $0.68$ & $0.87$ & $0.93$ & $1.03$
        & $0.79$ & $0.79$ & $0.91$ & $0.83$ & $1.00$ & $0.99$ \\
		& \textbf{\underline{H}}  
        & $1.07$ & $0.42$ & $0.32$ & $0.66$ & $0.58$ & $0.45$
        & $1.34$ & $1.03$ & $0.82$ & $0.92$ & $1.02$ & $1.33$
        & $0.83$ & $0.90$ & $0.92$ & $0.90$ & $1.00$ & $0.99$ \\
		& \textbf{\underline{F}} 
        & $1.33$ & $1.08$ & $1.24$ & $1.21$ & $1.14$ & $0.73$
        & $1.47$ & $1.19$ & $0.90$ & $1.10$ & $1.06$ & $1.07$
        & $0.75$ & $0.97$ & $1.00$ & $0.93$ & $0.96$ & $0.99$ \\
		& \textbf{\underline{U}}
        & $3.70$ & $1.55$ & $1.63$ & $0.70$ & $1.89$ & $0.69$
        & $1.73$ & $1.52$ & $1.47$ & $2.34$ & $2.77$ & $1.06$
        & $0.72$ & $0.84$ & $0.91$ & $0.88$ & $0.95$ & $0.99$ \\
		& \textbf{\underline{C}} 
        & $1.30$ & $1.11$ & $1.30$ & $0.73$ & $2.36$ & $0.79$
        & $3.59$ & $3.62$ & $2.68$ & $4.94$ & $1.62$ & $3.41$
        & $0.76$ & $0.95$ & $0.93$ & $0.90$ & $0.98$ & $1.00$ \\
        \midrule
        \multirow{5}{*}{RepE} 
		& \textbf{\underline{A}}
        & $1.94$ & $1.33$ & $1.32$ & - & - & -
        & $0.87$ & $1.22$ & $1.13$ & - & - & -
        & $0.81$ & $0.93$ & $0.95$ & - & - & - \\
		& \textbf{\underline{H}}
        & $1.18$ & $1.52$ & $1.68$ & - & - & -
        & $2.14$ & $2.10$ & $1.92$ & - & - & -
        & $0.80$ & $0.87$ & $0.93$ & - & - & - \\
		& \textbf{\underline{F}}
        & $1.86$ & $2.06$ & $1.90$ & - & - & -
        & $1.44$ & $1.50$ & $1.59$ & - & - & -
        & $0.93$ & $0.82$ & $0.94$ & - & - & - \\
		& \textbf{\underline{U}}
        & $2.08$ & $2.29$ & $0.58$ & - & - & -
        & $1.75$ & $2.02$ & $2.07$ & - & - & -
        & $0.97$ & $0.88$ & $0.72$ & - & - & - \\
		& \textbf{\underline{C}}
        & $1.26$ & $1.40$ & $1.39$ & - & - & -
        & $0.92$ & $0.76$ & $0.90$ & - & - & -
        & $0.95$ & $0.97$ & $0.67$ & - & - & - \\
        \bottomrule
        \end{tabular}
    }
    \end{adjustbox}
    \caption{Evaluation results after parameter selection for each model and attribute. Here, `Attr.' is short for `Attribute', Mean-MAE denotes the calibration error, standard deviation indicates the robustness of smooth control, and relevance suggests if the generated response aligns with the topic.  Some values are marked as `-' due to the constraints of accessing the model parameters and the coefficient range.}
  \label{tab:exp_result}%
\end{table*}

In this section, we apply our proposed evaluation framework, along with the constructed benchmark dataset, to assess the smooth control capability of various modern LLMs through two viable training-free methods: \textbf{(1)} Prompting with semantic shifters, and \textbf{(2) }Editing the internal model representations. We first introduce the experiment settings and then present the results and analyses.

\subsection{Baseline Methods For Smooth Control}

\noindent\textbf{Prompting LLMs.}
The most straightforward method to smoothly control the LLM to generate according to an attribute is to provide it with instruction on the degree level required. 
To achieve this, we need one description $\mathcal{D}_{\mathcal{A}, cv}$ for each degree $cv$ of the attribute $\mathcal{A}$:
\begin{equation*}
\begin{aligned}
    & \mathcal{R} = M_{\text{prompt}}(\mathcal{Q}, \mathcal{D}_{\mathcal{A}, cv}),
\end{aligned}
\end{equation*}
We call this prompting method parameterized by the descriptions we choose.
We consider two types of degree descriptions, first a list of semantic shifters that can describe the intensity paired with the adjective of the attribute (e.g., ``a little bit angry'' or ``very angry''), and the second, a crafted list of phrases that not necessary sticks with a format (e.g., ``slightly relaxed'' or ``extremely enraged''). 
The advantage of the first type is that they are seemingly easy to apply directly to different attributes, while for the second type, there is more flexibility in the descriptions.
The exact descriptions we use for each attribute and the prompt templates to use these descriptions are in Appendix~\ref{sec:template_prompting_degree_descriptions} and~\ref{sec:candidates}.

\paragraph{Representation Engineering (RepE).} Different from prompting, RepE~\cite{zou2023transparency} is a top-down approach to post-processing pretrained models via manipulating their internal representations for understanding and controlling neural networks. 

Specifically, it involves two distinct steps in particular. (1) \textbf{Reading:} localizing the functional representations for a specific concept, which is generally achieved by analyzing the neural activities after stimulating the model with certain input prompts. The original stimulus prompts are manually written by humans, which have limited scope and lack generalizability to unseen concepts. In our experiments, we employ GPT-4~\cite{openai2023gpt4} to generate those stimuli automatically and the prompt template is in Appendix~\ref{sec:stimulus}. (2) \textbf{Controlling.} The extracted representations from the reading step are then utilized as high-dimensional vectors to perturb the original model representations to different extents indicated by a control strength, which perfectly aligns with the concept of control value in our task. Therefore, we specify the control strength for each control value of the attribute A. Such manipulation of the internal representations is also parameterized by the strength we indicate.
\begin{equation*}
\begin{aligned}
    & \mathcal{R} = M_{\text{RepE}}(\mathcal{Q}, \text{Strength}_{\mathcal{A}, cv}),
\end{aligned}
\end{equation*}

\subsection{Parameter Selection}
\label{sec:calibration_op}
It is not immediately clear whether the human-interpreted degree descriptions for prompting or the human-selected degree intensities in RepE transfer to a smooth degree control for the LLM. 
Therefore, we consider a ``parameter selection'' process for these two methods for calibration of the degrees.
Specifically, we proactively consider a larger number of degree parameters (descriptions for prompting or strength for RepE), and obtain generations of the LLM based on the parameter through a held-out set of questions.
Then, we select the sequence of parameters that leads to the best overall metric, which is defined and calculated as: 

\vspace{-1em}
\begin{equation*}
\begin{aligned}
    & \text{Metric}= \frac{\text{Mean-MAE} + \text{Mean-STD}}{(R_{\text{max}} -R_{\text{min}})*\text{Relevance}}
\end{aligned}
\end{equation*}

This metric is designed to determine a better set of generations from a specific smooth control method. The breakdown and intuitive explanations of this formula are as follows. \textbf{(1)} The nominator is the sum of the two aforementioned rating errors. A high Mean-MAE indicates misalignment with rating scales, while a high Mean-STD indicates unstable, varied ratings. To keep both values reasonable, we add rather than multiply them, as they share the same scale. Empirical evaluation shows that an unweighted average performs nearly best based on human inspection. The corresponding statistics are exhibited in Appendix~\ref{sec:ps_analysis}. \textbf{(2)} The denominator is the multiplication of the normalization term and the relevance penalty factor. A low relevance score is undesirable, so we use its reciprocal to heavily penalize low-relevance generations.

The selection can be done efficiently by brute-force enumeration when the number of the total considered parameters is not too large and the specific number in our case is 20.

\subsection{Experiment Settings}
The evaluations are conducted on diverse LLMs for the smooth control of specific attributes. As such, we present the models, attributes, and datasets that are utilized in the experiments here.

\paragraph{Models.} We employ both open-source and closed-source LLMs for our experiments. Specifically, we adopt Mistral~\citep{jiang2023mistral} and LLaMA2~\citep{touvron2023llama} at different scales for the experiments of editing the internal model representations, as it requires access to the model parameters. For prompting with semantic shifters, we further utilize GPT-3.5~\citep{openai2022chatgpt} and GPT-4~\citep{openai2023gpt4} models.

\paragraph{Attribute.} As explained in Section~\ref{sec:benchmark}, we select \anger, \happiness, \formality, \understandability, and \conciseness as the attributes to evaluate. In particular, the intensity of \conciseness is measured differently than other attributes by directly counting the number of words in the responses.

\paragraph{Dataset.} We adopt the constructed benchmark dataset introduced in Section~\ref{sec:benchmark}, which consists of $1,500$ query sentences in total, with $500$ for each of the aforementioned 5 attributes.

\paragraph{Metric.} According to our evaluation framework introduced in Section~\ref{sec:framework}, we adopt mean-MAE, standard deviation, and relevance as the main metrics.

\section{Experiment results}

\begin{figure*}[ht]
    \centering
    \includegraphics[width=0.95\linewidth]{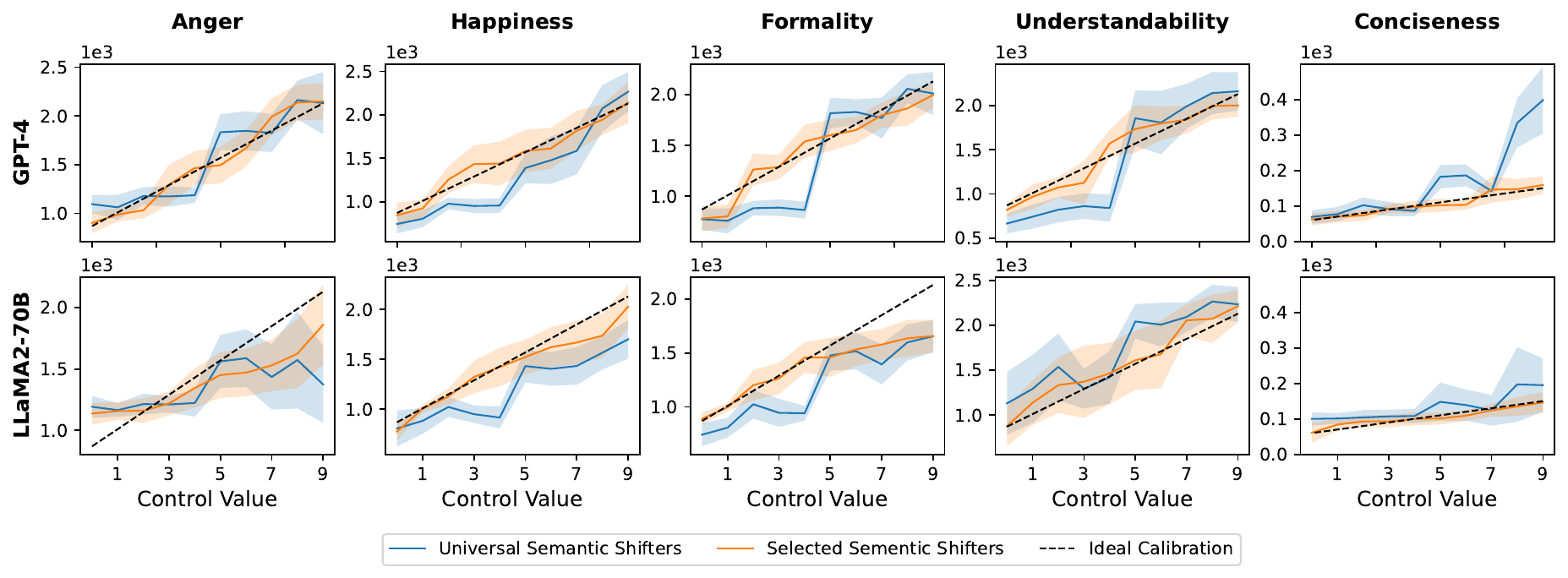}
    \caption{Comparisons between prompting with universal and selected semantic shifters. The Y axis is the attribute intensity. The black dashed lines are the ideal correlation between the control value and the attribute intensity. }
    \label{fig:universal_baseline}
\end{figure*}

\subsection{Main Results}
Table~\ref{tab:exp_result} shows the smooth control performance achieved by different models with different methods, on several attributes.
One can observe that GPT-4 is significantly better than other models for all attributes, especially in terms of Mean-MAE, namely the consistency between the control values and the obtained attribute intensities. GPT-4 is also significantly better than other models in terms of the relevance between the model's response and the query, despite the potential cause being that GPT-4 is also used to evaluate such relevance. 

Interestingly, we observe that model sizes may negatively affect the smooth performance. A relatively fair test bed for this is the Llama family, where one can observe that for most attributes, Mean-MAE decreases constantly as the model size increases from 7B, 13B, to 70B.

Finally, we also observe that prompting is almost as good as, if relatively better than repE. This implies that prompting is preferred in realistic applications of smooth control since it requires no access to the internal model representations and thus can be potentially applied to more LLMs.

\begin{table*}[ht]
  \centering
   \small
    \begin{adjustbox}{max width=1\linewidth}
    {
        \begin{tabular}{c|cccccc|cccccc|cccccc}
        \toprule
        & \multicolumn{6}{c|}{\textbf{Mean-MAE ($\downarrow$)}}  
        & \multicolumn{6}{c}{\textbf{Mean-STD ($\downarrow$)}}  
        & \multicolumn{6}{c}{\textbf{Relevance  ($\uparrow$)}} \\
        & \textbf{mistral} & \textbf{llama} & \textbf{llama} & \textbf{llama} & \textbf{gpt} & \textbf{gpt} & \textbf{mistral} & \textbf{llama} & \textbf{llama} & \textbf{llama} & \textbf{gpt} & \textbf{gpt} & \textbf{mistral} & \textbf{llama-7b} & \textbf{llama} & \textbf{llama} & \textbf{gpt} & \textbf{gpt}
        \\
        & \textbf{7b} & \textbf{7b} & \textbf{13b} & \textbf{70b} & \textbf{3.5} & \textbf{4} & \textbf{7b} & \textbf{7b} & \textbf{13b} & \textbf{70b} & \textbf{3.5} & \textbf{4} & \textbf{7b} & \textbf{7b} & \textbf{13b} & \textbf{70b} & \textbf{3.5} & \textbf{4} \\
        \midrule 
	\textbf{\underline{A}} 
        & $1.94$ & $1.08$ & $0.84$ & $1.49$ & $0.86$ & $0.85$ & $1.17$ & $1.10$ & $1.20$ & $1.67$ & $1.23$ & $1.47$ & $0.78$ & $0.75$ & $0.86$ & $0.72$ & $0.92$ & $0.99$ \\
	\textbf{\underline{H}}  & $1.26$ & $0.51$ & $0.50$ & $0.96$ & $0.77$ & $1.25$ & $1.48$ & $1.25$ & $1.23$ & $1.23$ & $1.42$ & $1.60$ & $0.81$ & $0.87$ & $0.94$ & $0.87$ & $1.00$ & $0.99$ \\
	\textbf{\underline{F}}  & $1.46$ & $1.48$ & $1.54$ & $1.50$ & $1.36$ & $1.29$ & $1.19$ & $1.23$ & $1.03$ & $1.23$ & $1.16$ & $1.12$ & $0.75$ & $0.95$ & $0.99$ & $0.91$ & $0.96$ & $0.99$ \\
	\textbf{\underline{U}}  & $3.35$ & $1.65$ & $1.70$ & $2.68$ & $4.03$ & $2.30$ & $1.67$ & $1.56$ & $1.63$ & $2.08$ & $2.09$ & $1.12$ & $0.45$ & $0.81$ & $0.83$ & $0.91$ & $0.86$ & $0.93$ \\
	\textbf{\underline{C}}  & $1.60$ & $1.08$ & $1.30$ & $0.90$ & $3.19$ & $1.63$ & $2.66$ & $2.71$ & $1.98$ & $2.43$ & $1.36$ & $2.13$ & $0.77$ & $0.92$ & $0.93$ & $0.91$ & $0.89$ & $1.00$ \\
        \bottomrule
        \end{tabular}
    }
    \end{adjustbox}
    \caption{
    Prompting with intensity descriptors that are not specific to models.
    }
  \label{tab:exp-unknown-model}%
\end{table*}

\subsection{Specificity of Parameter Selection in Intensity Calibration}

In this section, we wish to explore whether the intensity descriptors we selected for each attribute and each model are specific to the model or the attribute. Here we mainly conduct the investigation based on prompting since prompting is more preferred than RepE based on the above results.

\begin{table}[h]
    \small
    \centering
    \begin{tabularx}{0.6\linewidth}{c|c}
\toprule
extremely not & a little bit\\
very not & somewhat\\
moderately not & moderately\\
somewhat not & very\\
a little bit not & extremely \\
\bottomrule
\end{tabularx}
    \caption{Universal Semantic Shifters}
    \vspace{-1em}
    \label{tab:universal_semantic_shifters}
\end{table}

\paragraph{Should intensity descriptors be specific to attribute?}
We validate whether it is possible to use a universal set of descriptors to control the intensities of all attributes listed. If possible, such a set can greatly ease the implementation of smooth control of LLMs and reduce the inference cost for selecting specific descriptors of each attribute.

To this end, we experiment with using a set of fixed semantic shifters to modulate the intensity of an attribute in prompting. In specific, we prompt GPT-4 multiple times to generate $30$ different adverbs of degrees that are commonly used. We then select $10$ that appear the most frequently in responses, which are shown in Table~\ref{tab:universal_semantic_shifters}.

Figure~\ref{fig:universal_baseline} and Table~\ref{tab:exp_baseline} show the results of using fixed semantic shifters for prompting LLMs. We observe that such fixed semantic shifters achieve significantly worse performance in smooth control, especially in terms of Mean-MAE. This means fixed semantic shifters cannot properly control the attribute to match the desired intensities.

\paragraph{Are intensity descriptors specific to model?}
Among our experiment results, we found that the intensity descriptors selected to achieve the best smooth control performance vary significantly across models. According to our observation of the results for attribute ``Formality'', to achieve an intensity level of $3$, GPT-4 would prefer ``Highly Inappropriate'' in its prompt. In contrast, Llama2-70b would prefer ``Neural'' to achieve the same intensity level. Further, we found that the intensity descriptors preferred by different models may not even be consistent in terms of order.

\begin{table}[t]
  \centering
   \small
    \begin{adjustbox}{max width=1\linewidth}
    {
        \begin{tabular}{l|cc|cc|cc}
        \toprule
              & \multicolumn{2}{c|}{\textbf{Mean-MAE}} & \multicolumn{2}{c|}{\textbf{Mean-STD}}& \multicolumn{2}{c}{\textbf{Relevance}} \\
        \textbf{Attri} & \textbf{gpt-4} & \textbf{llama-70b} & \textbf{gpt-4} & \textbf{llama-70b} & \textbf{gpt-4} & \textbf{llama-70B}\\
        \midrule
        \textbf{\underline{A}}             & $1.00$ & $2.02$  & $1.27$ & $1.45$ & $0.98$ & $0.77$ \\
        \textbf{\underline{H}}          & $1.76$ & $2.29$  & $1.31$ & $1.18$ & $0.97$ & $0.83$ \\
        \textbf{\underline{F}}          & $1.75$ & $2.3$  & $1.03$ & $1.06$ & $0.99$ & $0.95$ \\
        \textbf{\underline{U}}      & $2.01$ & $1.84$  & $1.64$ & $2.07$ & $1.00$ & $0.83$ \\
        \textbf{\underline{C}} & $7.17$ & $3.07$ & $3.81$ & $4.61$ & $1.00$ & $0.86$ \\
        \bottomrule
        \end{tabular}
    }
    \end{adjustbox}
    \caption{Baseline with fixed semantic shifters.}
    \vspace{-1.5em}
  \label{tab:exp_baseline}%
\end{table}

Since the best intensity descriptors are not specific to the model, one cannot simply transfer the intensity descriptors selected for one model to another model. We conduct an additional experiment to demonstrate this. In Table~\ref{tab:exp-unknown-model}, for each model, we use all models except this model to select the intensity descriptors. One may observe that these intensity descriptors transferred from other models cause significantly worse smooth control performance, especially in terms of Mean-MAE. This shows that it is necessary to select intensity descriptors specific to each model to properly control the attribute to match the desired intensity.

\section{Conclusions and Future Work}
This work studies smoothly controllable text generation for large language models.
We created an evaluation system on five different attributes to evaluate smooth control methods on different intensity levels for three metrics: error, variation of the generated sentence's intensities and the relevance to the generation questions.
The system is implemented based on Elo ratings, automatically evaluating using LLMs, and designed to be efficient in evaluation.
We evaluate two representative methods, prompting and representation engineering.
We find that (1) Model sizes may negatively affect the smooth performance. (2) Prompting is almost as good as, if relatively better than repE.

\section*{Limitations}
Our work presents an evaluation of smooth control methods for LLM generations.
There are several limitations that we have considered:
\begin{itemize}[nosep,leftmargin=*]
    \item We used GPT-4 as an automatic evaluator in building our evaluation system, mainly for reducing human effort. While we have verified its closeness with human preference on all the 5 attributes we considered, we admit that our system will suffer from the same limitations of using LLM as annotators, such as not being robust to certain (manually crafted) sentences, and not being a free service to use, especially that we find not as competent LLMs (e.g., GPT-3.5) do not have a similar strong annotation power.
    \item We mainly evaluated two training-free methods, Prompting and Representation Engineering for their soft control ability, due to their simplicity and representativeness. Other soft control methods, including some that require model fine-tuning, could be evaluated in future work. 
\end{itemize}

\section*{Acknowledgement}
We thank the anonymous reviewers for their insightful comments. Our work is sponsored in part by NSF CAREER Award 2239440, NSF Proto-OKN Award 2333790, as well as generous gifts from Google, Adobe, and Teradata. Any opinions, findings, conclusions, or recommendations expressed herein are those of the authors and should not be interpreted as necessarily representing the views, either expressed or implied, of the U.S. Government. The U.S. Government is authorized to reproduce and distribute reprints for government purposes notwithstanding any copyright annotation hereon.

\bibliography{custom}

\begin{thebibliography}{38}
\expandafter\ifx\csname natexlab\endcsname\relax\def\natexlab#1{#1}\fi

\bibitem[{Dathathri et~al.(2019)Dathathri, Madotto, Lan, Hung, Frank, Molino, Yosinski, and Liu}]{Dathathri2019PlugAP}
Sumanth Dathathri, Andrea Madotto, Janice Lan, Jane Hung, Eric Frank, Piero Molino, Jason Yosinski, and Rosanne Liu. 2019.
\newblock \href {https://api.semanticscholar.org/CorpusID:208617790} {Plug and play language models: A simple approach to controlled text generation}.
\newblock \emph{ArXiv}, abs/1912.02164.

\bibitem[{Hu et~al.(2017)Hu, Yang, Liang, Salakhutdinov, and Xing}]{Hu2017TowardCG}
Zhiting Hu, Zichao Yang, Xiaodan Liang, Ruslan Salakhutdinov, and Eric~P. Xing. 2017.
\newblock \href {https://api.semanticscholar.org/CorpusID:20981275} {Toward controlled generation of text}.
\newblock In \emph{International Conference on Machine Learning}.

\bibitem[{Huang et~al.(2018)Huang, Zaiane, Trabelsi, and Dziri}]{Huang2018AutomaticDG}
Chenyang Huang, Osmar~R Zaiane, Amine Trabelsi, and Nouha Dziri. 2018.
\newblock \href {https://api.semanticscholar.org/CorpusID:13788863} {Automatic dialogue generation with expressed emotions}.
\newblock In \emph{North American Chapter of the Association for Computational Linguistics}.

\bibitem[{Huang et~al.(2022)Huang, Zhang, Ko, Liu, Wu, Wang, and Tang}]{Huang2022PersonalizedDG}
Qiushi Huang, Yu~Zhang, Tom Ko, Xubo Liu, Boyong Wu, Wenwu Wang, and Lilian Tang. 2022.
\newblock \href {https://api.semanticscholar.org/CorpusID:253157530} {Personalized dialogue generation with persona-adaptive attention}.
\newblock \emph{ArXiv}, abs/2210.15088.

\bibitem[{Jiang et~al.(2023)Jiang, Sablayrolles, Mensch, Bamford, Chaplot, Casas, Bressand, Lengyel, Lample, Saulnier et~al.}]{jiang2023mistral}
Albert~Q Jiang, Alexandre Sablayrolles, Arthur Mensch, Chris Bamford, Devendra~Singh Chaplot, Diego de~las Casas, Florian Bressand, Gianna Lengyel, Guillaume Lample, Lucile Saulnier, et~al. 2023.
\newblock \href {https://arxiv.org/abs/2310.06825} {Mistral 7b}.
\newblock \emph{ArXiv preprint}, abs/2310.06825.

\bibitem[{Jin et~al.(2020)Jin, Jin, Zhou, Orii, and Szolovits}]{Jin2020HooksIT}
Di~Jin, Zhijing Jin, Joey~Tianyi Zhou, Lisa Orii, and Peter Szolovits. 2020.
\newblock \href {https://api.semanticscholar.org/CorpusID:214802410} {Hooks in the headline: Learning to generate headlines with controlled styles}.
\newblock In \emph{Annual Meeting of the Association for Computational Linguistics}.

\bibitem[{Jin et~al.(2019)Jin, Jin, Mueller, Matthews, and Santus}]{Jin2019IMaTUT}
Zhijing Jin, Di~Jin, Jonas~W. Mueller, Nicholas Matthews, and Enrico Santus. 2019.
\newblock \href {https://api.semanticscholar.org/CorpusID:202541632} {Imat: Unsupervised text attribute transfer via iterative matching and translation}.
\newblock In \emph{Conference on Empirical Methods in Natural Language Processing}.

\bibitem[{Khalifa et~al.(2020)Khalifa, ElSahar, and Dymetman}]{Khalifa2020ADA}
Muhammad Khalifa, Hady ElSahar, and Marc Dymetman. 2020.
\newblock \href {https://api.semanticscholar.org/CorpusID:229348988} {A distributional approach to controlled text generation}.
\newblock \emph{ArXiv}, abs/2012.11635.

\bibitem[{Krause et~al.(2020)Krause, Gotmare, McCann, Keskar, Joty, Socher, and Rajani}]{Krause2020GeDiGD}
Ben Krause, Akhilesh~Deepak Gotmare, Bryan McCann, Nitish~Shirish Keskar, Shafiq~R. Joty, Richard Socher, and Nazneen Rajani. 2020.
\newblock \href {https://api.semanticscholar.org/CorpusID:221655075} {Gedi: Generative discriminator guided sequence generation}.
\newblock In \emph{Conference on Empirical Methods in Natural Language Processing}.

\bibitem[{Lester et~al.(2021)Lester, Al-Rfou, and Constant}]{Lester2021ThePO}
Brian Lester, Rami Al-Rfou, and Noah Constant. 2021.
\newblock \href {https://api.semanticscholar.org/CorpusID:233296808} {The power of scale for parameter-efficient prompt tuning}.
\newblock In \emph{Conference on Empirical Methods in Natural Language Processing}.

\bibitem[{Li et~al.(2018)Li, Jia, He, and Liang}]{Li2018DeleteRG}
Juncen Li, Robin Jia, He~He, and Percy Liang. 2018.
\newblock \href {https://api.semanticscholar.org/CorpusID:4937880} {Delete, retrieve, generate: a simple approach to sentiment and style transfer}.
\newblock In \emph{North American Chapter of the Association for Computational Linguistics}.

\bibitem[{Li and Liang(2021)}]{Li2021PrefixTuningOC}
Xiang~Lisa Li and Percy Liang. 2021.
\newblock \href {https://api.semanticscholar.org/CorpusID:230433941} {Prefix-tuning: Optimizing continuous prompts for generation}.
\newblock \emph{Proceedings of the 59th Annual Meeting of the Association for Computational Linguistics and the 11th International Joint Conference on Natural Language Processing (Volume 1: Long Papers)}, abs/2101.00190.

\bibitem[{Liu et~al.(2020{\natexlab{a}})Liu, Chen, Chen, Lou, Chen, Zhou, and Zhang}]{Liu2020YouIM}
Qian Liu, Yihong Chen, B.~Chen, Jian-Guang Lou, Zixuan Chen, Bin Zhou, and Dongmei Zhang. 2020{\natexlab{a}}.
\newblock \href {https://api.semanticscholar.org/CorpusID:215745354} {You impress me: Dialogue generation via mutual persona perception}.
\newblock In \emph{Annual Meeting of the Association for Computational Linguistics}.

\bibitem[{Liu et~al.(2020{\natexlab{b}})Liu, Xu, Jia, Ma, Wang, and Vosoughi}]{Liu2020DataBT}
Ruibo Liu, Guangxuan Xu, Chenyan Jia, Weicheng Ma, Lili Wang, and Soroush Vosoughi. 2020{\natexlab{b}}.
\newblock \href {https://api.semanticscholar.org/CorpusID:226262374} {Data boost: Text data augmentation through reinforcement learning guided conditional generation}.
\newblock In \emph{Conference on Empirical Methods in Natural Language Processing}.

\bibitem[{Ma et~al.(2020)Ma, Sap, Rashkin, and Choi}]{Ma2020PowerTransformerUC}
Xinyao Ma, Maarten Sap, Hannah Rashkin, and Yejin Choi. 2020.
\newblock \href {https://api.semanticscholar.org/CorpusID:225075985} {Powertransformer: Unsupervised controllable revision for biased language correction}.
\newblock In \emph{Conference on Empirical Methods in Natural Language Processing}.

\bibitem[{Madotto et~al.(2020)Madotto, Lin, Bang, and Fung}]{Madotto2020TheAA}
Andrea Madotto, Zhaojiang Lin, Yejin Bang, and Pascale Fung. 2020.
\newblock \href {https://api.semanticscholar.org/CorpusID:221370525} {The adapter-bot: All-in-one controllable conversational model}.
\newblock In \emph{AAAI Conference on Artificial Intelligence}.

\bibitem[{Niu and Bansal(2018)}]{Niu2018PoliteDG}
Tong Niu and Mohit Bansal. 2018.
\newblock \href {https://api.semanticscholar.org/CorpusID:13690180} {Polite dialogue generation without parallel data}.
\newblock \emph{Transactions of the Association for Computational Linguistics}, 6:373--389.

\bibitem[{OpenAI(2022)}]{openai2022chatgpt}
OpenAI. 2022.
\newblock Introducing chatgpt.
\newblock \url{https://openai.com/blog/chatgpt}.

\bibitem[{OpenAI(2023)}]{openai2023gpt4}
OpenAI. 2023.
\newblock \href {http://arxiv.org/abs/2303.08774} {Gpt-4 technical report}.

\bibitem[{Pryzant et~al.(2019)Pryzant, Martinez, Dass, Kurohashi, Jurafsky, and Yang}]{Pryzant2019AutomaticallyNS}
Reid Pryzant, Richard~Diehl Martinez, Nathan Dass, Sadao Kurohashi, Dan Jurafsky, and Diyi Yang. 2019.
\newblock \href {https://api.semanticscholar.org/CorpusID:208248333} {Automatically neutralizing subjective bias in text}.
\newblock \emph{ArXiv}, abs/1911.09709.

\bibitem[{Rao and Tetreault(2018)}]{Rao2018DearSO}
Sudha Rao and Joel~R. Tetreault. 2018.
\newblock \href {https://api.semanticscholar.org/CorpusID:4859003} {Dear sir or madam, may i introduce the gyafc dataset: Corpus, benchmarks and metrics for formality style transfer}.
\newblock In \emph{North American Chapter of the Association for Computational Linguistics}.

\bibitem[{Ribeiro et~al.(2023)Ribeiro, Bansal, and Dreyer}]{Ribeiro2023GeneratingSW}
Leonardo F.~R. Ribeiro, Mohit Bansal, and Markus Dreyer. 2023.
\newblock \href {https://api.semanticscholar.org/CorpusID:264172439} {Generating summaries with controllable readability levels}.
\newblock In \emph{Conference on Empirical Methods in Natural Language Processing}.

\bibitem[{Ribeiro et~al.(2021)Ribeiro, Zhang, and Gurevych}]{Ribeiro2021StructuralAI}
Leonardo F.~R. Ribeiro, Yue Zhang, and Iryna Gurevych. 2021.
\newblock \href {https://api.semanticscholar.org/CorpusID:232240435} {Structural adapters in pretrained language models for amr-to-text generation}.
\newblock \emph{ArXiv}, abs/2103.09120.

\bibitem[{Shen et~al.(2017)Shen, Lei, Barzilay, and Jaakkola}]{Shen2017StyleTF}
Tianxiao Shen, Tao Lei, Regina Barzilay, and T.~Jaakkola. 2017.
\newblock \href {https://api.semanticscholar.org/CorpusID:7296803} {Style transfer from non-parallel text by cross-alignment}.
\newblock \emph{ArXiv}, abs/1705.09655.

\bibitem[{Song et~al.(2021)Song, Wang, Zhang, Zhang, and Liu}]{Song2021BoBBO}
Haoyu Song, Yan Wang, Kaiyan Zhang, Weinan Zhang, and Ting Liu. 2021.
\newblock \href {https://api.semanticscholar.org/CorpusID:235417177} {Bob: Bert over bert for training persona-based dialogue models from limited personalized data}.
\newblock \emph{ArXiv}, abs/2106.06169.

\bibitem[{Song et~al.(2019)Song, Zheng, Liu, Xu, and Huang}]{Song2019GeneratingRW}
Zhenqiao Song, Xiaoqing Zheng, Lu~Liu, Mu~Xu, and Xuanjing Huang. 2019.
\newblock \href {https://api.semanticscholar.org/CorpusID:196193055} {Generating responses with a specific emotion in dialog}.
\newblock In \emph{Annual Meeting of the Association for Computational Linguistics}.

\bibitem[{Tambwekar et~al.(2018)Tambwekar, Dhuliawala, Martin, Mehta, Harrison, and Riedl}]{Tambwekar2018ControllableNS}
Pradyumna Tambwekar, Murtaza Dhuliawala, Lara~J. Martin, Animesh Mehta, Brent Harrison, and Mark~O. Riedl. 2018.
\newblock \href {https://api.semanticscholar.org/CorpusID:199465680} {Controllable neural story plot generation via reward shaping}.
\newblock In \emph{International Joint Conference on Artificial Intelligence}.

\bibitem[{Touvron et~al.(2023)Touvron, Martin, Stone, Albert, Almahairi, Babaei, Bashlykov, Batra, Bhargava, Bhosale et~al.}]{touvron2023llama}
Hugo Touvron, Louis Martin, Kevin Stone, Peter Albert, Amjad Almahairi, Yasmine Babaei, Nikolay Bashlykov, Soumya Batra, Prajjwal Bhargava, Shruti Bhosale, et~al. 2023.
\newblock \href {https://arxiv.org/abs/2307.09288} {Llama 2: Open foundation and fine-tuned chat models}.
\newblock \emph{ArXiv preprint}, abs/2307.09288.

\bibitem[{Wolf et~al.(2019)Wolf, Sanh, Chaumond, and Delangue}]{Wolf2019TransferTransfoAT}
Thomas Wolf, Victor Sanh, Julien Chaumond, and Clement Delangue. 2019.
\newblock \href {https://api.semanticscholar.org/CorpusID:59222757} {Transfertransfo: A transfer learning approach for neural network based conversational agents}.
\newblock \emph{ArXiv}, abs/1901.08149.

\bibitem[{Xu et~al.(2020)Xu, Patwary, Shoeybi, Puri, Fung, Anandkumar, and Catanzaro}]{Xu2020ControllableSG}
Peng Xu, Mostofa Patwary, Mohammad Shoeybi, Raul Puri, Pascale Fung, Anima Anandkumar, and Bryan Catanzaro. 2020.
\newblock \href {https://api.semanticscholar.org/CorpusID:222125036} {Controllable story generation with external knowledge using large-scale language models}.
\newblock In \emph{Conference on Empirical Methods in Natural Language Processing}.

\bibitem[{Yang et~al.(2022)Yang, Liu, Lei, Yang, Xue, Chen, and Xie}]{Yang2022TailorAP}
Kexin Yang, Dayiheng Liu, Wenqiang Lei, Baosong Yang, Mingfeng Xue, Boxing Chen, and Jun Xie. 2022.
\newblock \href {https://api.semanticscholar.org/CorpusID:248426828} {Tailor: A prompt-based approach to attribute-based controlled text generation}.
\newblock \emph{ArXiv}, abs/2204.13362.

\bibitem[{Zeldes et~al.(2020)Zeldes, Padnos, Sharir, and Peleg}]{Zeldes2020TechnicalRA}
Yoel Zeldes, Dan Padnos, Or~Sharir, and Barak Peleg. 2020.
\newblock \href {https://api.semanticscholar.org/CorpusID:220265537} {Technical report: Auxiliary tuning and its application to conditional text generation}.
\newblock \emph{ArXiv}, abs/2006.16823.

\bibitem[{Zhang et~al.(2020)Zhang, Wang, Li, Gan, Brockett, and Dolan}]{Zhang2020POINTERCP}
Yizhe Zhang, Guoyin Wang, Chunyuan Li, Zhe Gan, Chris Brockett, and Bill Dolan. 2020.
\newblock \href {https://api.semanticscholar.org/CorpusID:226604173} {Pointer: Constrained progressive text generation via insertion-based generative pre-training}.
\newblock In \emph{Conference on Empirical Methods in Natural Language Processing}.

\bibitem[{Zhang et~al.(2018)Zhang, Ren, Liu, Wang, Chen, Li, Zhou, and Chen}]{Zhang2018StyleTA}
Zhirui Zhang, Shuo Ren, Shujie Liu, Jianyong Wang, Peng Chen, Mu~Li, Ming Zhou, and Enhong Chen. 2018.
\newblock \href {https://api.semanticscholar.org/CorpusID:266349854} {Style transfer as unsupervised machine translation}.
\newblock \emph{ArXiv}, abs/1808.07894.

\bibitem[{Zheng et~al.(2023)Zheng, Chiang, Sheng, Zhuang, Wu, Zhuang, Lin, Li, Li, Xing, Zhang, Gonzalez, and Stoica}]{zheng2023judging}
Lianmin Zheng, Wei-Lin Chiang, Ying Sheng, Siyuan Zhuang, Zhanghao Wu, Yonghao Zhuang, Zi~Lin, Zhuohan Li, Dacheng Li, Eric.~P Xing, Hao Zhang, Joseph~E. Gonzalez, and Ion Stoica. 2023.
\newblock \href {http://arxiv.org/abs/2306.05685} {Judging llm-as-a-judge with mt-bench and chatbot arena}.

\bibitem[{Zheng et~al.(2019)Zheng, Zhang, Mao, and Huang}]{Zheng2019APB}
Yinhe Zheng, Rongsheng Zhang, Xiaoxi Mao, and Minlie Huang. 2019.
\newblock \href {https://api.semanticscholar.org/CorpusID:207863734} {A pre-training based personalized dialogue generation model with persona-sparse data}.
\newblock In \emph{AAAI Conference on Artificial Intelligence}.

\bibitem[{Ziegler et~al.(2019)Ziegler, Stiennon, Wu, Brown, Radford, Amodei, Christiano, and Irving}]{Ziegler2019FineTuningLM}
Daniel~M. Ziegler, Nisan Stiennon, Jeff Wu, Tom~B. Brown, Alec Radford, Dario Amodei, Paul Christiano, and Geoffrey Irving. 2019.
\newblock \href {https://api.semanticscholar.org/CorpusID:202660943} {Fine-tuning language models from human preferences}.
\newblock \emph{ArXiv}, abs/1909.08593.

\bibitem[{Zou et~al.(2023)Zou, Phan, Chen, Campbell, Guo, Ren, Pan, Yin, Mazeika, Dombrowski et~al.}]{zou2023transparency}
Andy Zou, Long Phan, Sarah Chen, James Campbell, Phillip Guo, Richard Ren, Alexander Pan, Xuwang Yin, Mantas Mazeika, Ann-Kathrin Dombrowski, et~al. 2023.
\newblock Representation engineering: A top-down approach to ai transparency.
\newblock \emph{arXiv preprint arXiv:2310.01405}.

\end{thebibliography}

\appendix
\clearpage
\section*{Appendices}

\section{Prompt Templates}
We list all prompt templates we used in this paper.

\subsection{Question Generation}
\label{sec:template_question_generation}
Our dataset consists of questions that potentially can be answered with different degrees of attributes. The template to generate the questions is
\begin{lstlisting}[language=Python, numbers=none]
Generate 10 prompts that can be answered with varying degrees of <concept>.
\end{lstlisting}

\subsection{Pairwise Annotation}
\label{sec:template_pairwise_annotation}
This template is used to compare two responses to decide which shows a greater degree of the concept.
\begin{lstlisting}[language=Python, numbers=none]
For each pair of responses, identify which response expresses more <concept>. Write the pair number followed by '1' if the first response is more <concept>, or '2' if the second response is more <concept>. Format your response like this: '1. 1', '2. 2', etc.
\end{lstlisting}

\subsection{Relevance Annotation}
\label{sec:template_relevance_annotation}
This template is used to Judge if a response is relevant (1) or not (0) to the query.
\begin{lstlisting}[language=Python, numbers=none]
Given the following query and response, please assess whether the response is relevant to the query. Answer with '1' if the response is relevant, and '0' if it is not relevant.
\end{lstlisting}

\subsection{Prompting with Degree Descriptions}
\label{sec:template_prompting_degree_descriptions}
This template is used to respond to queries with a specified emotional tone or style.
\begin{lstlisting}[language=Python, numbers=none]
Please respond to {{queries[i]}} with a paragraph in a [tone | style] that is {{semantic shifter}}. The response should be three sentences long.
\end{lstlisting}

\subsection{Generating Degree Descriptions}
\label{sec:template_generate_degree_descriptions}
This template is used to identify words or phrases that can shift the meaning of a concept, either intensifying or diminishing its strength.
\begin{lstlisting}[language=Python, numbers=none]
Describing <concept> levels on a scale from -9 to 10 using phrases.
\end{lstlisting}

\subsection{Stimulus Prompts Generation.}
\begin{lstlisting}[language=Python, numbers=none]
Generate 10 prompts that can stimulate <concept>.\end{lstlisting}
\label{sec:stimulus}

\subsection{Candidates for Semantic Shifters}
\label{sec:candidates}

\section{Parameter Selection Analysis}
\label{sec:ps_analysis}
We considered different sets of $\alpha$ (from 0 to 1) for the weighted average of Mean-MAE and Mean-STD to calculate the overall metric.
\begin{equation}
    \alpha \times \text{Mean-MAE} + (1-\alpha) \times \text{Mean-STD}
\end{equation}

\begin{figure}[t]
    \centering
    \includegraphics[width=\linewidth]{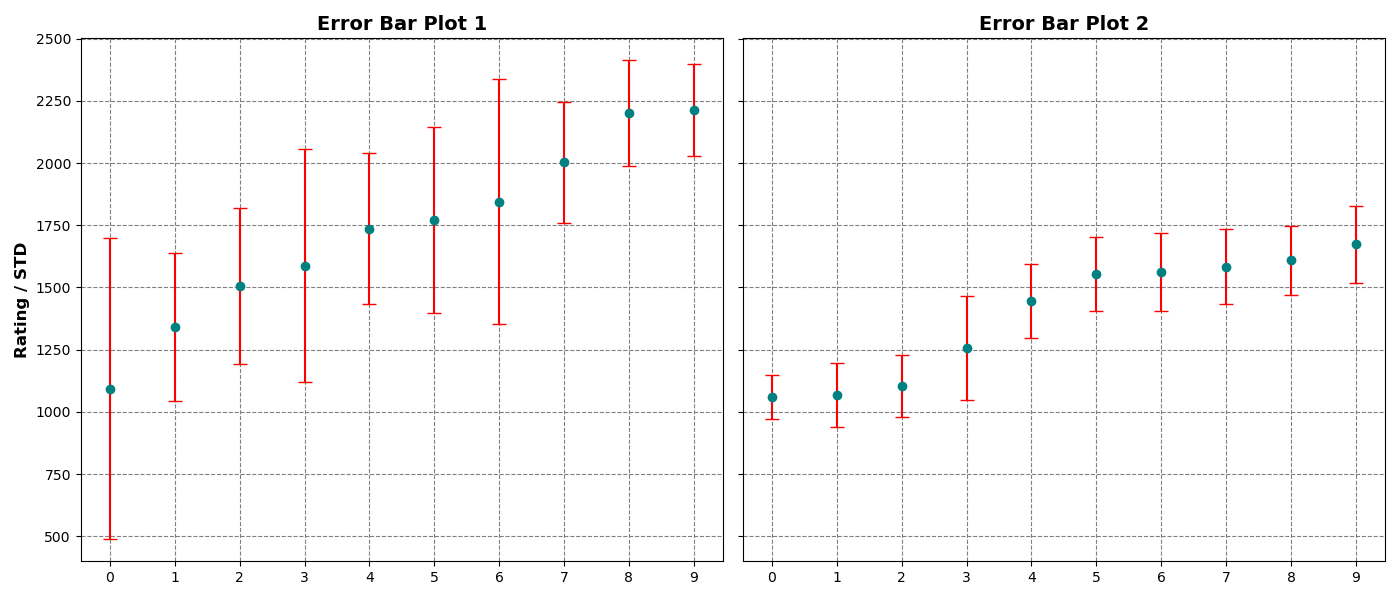}
    \caption{Examples for human evaluation.}
    
    \label{fig:eval_example}
\end{figure}

\begin{figure}[t]
    \centering
    \includegraphics[width=\linewidth]{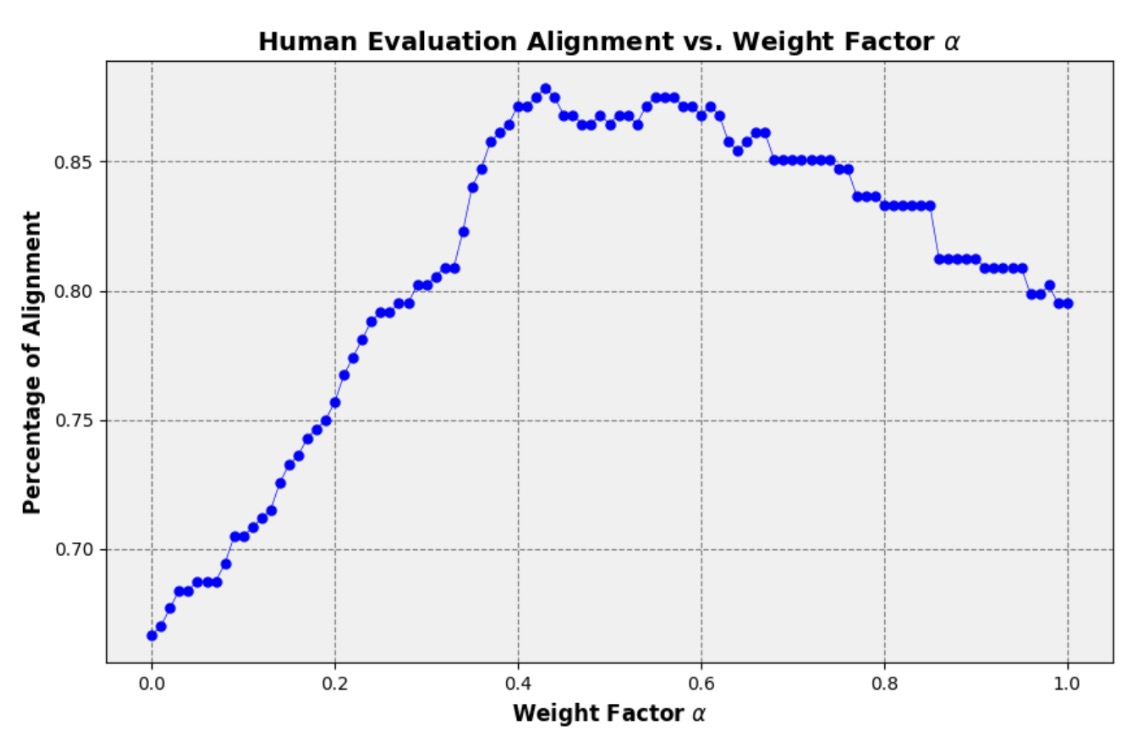}
    \caption{Alignment with humans for different weight factors.}
    
    \label{fig:alpha}
\end{figure}

For each weight factor $\alpha$, we considered pairs of error bar plots of the average and standard deviation values, and asked humans to judge which plot is better as shown in Figure~\ref{fig:eval_example}. We compare the human evaluation result with the result that our metric provides, and record the percentage of alignment. As shown in Figure~\ref{fig:alpha}, alignment follows a bell curve, peaking at 0.87 when $\alpha$ is 0.5-ish. Therefore, we directly adopt the vanilla average of these two rating errors rather than the weighted ones.

\section{Generated Data Examples}
\label{sec:data_example}

\begin{table*}[ht]
    \small
    \centering
    \begin{tabularx}{0.9\linewidth}{lX}
        \toprule
        \textbf{Attribute} & \textbf{Query} \\
        \midrule
        Anger & Why did you use my personal items without asking for permission? \\
          & What exactly is causing the continuous delays on the construction project you assured would be finished by now? \\
          & What possessed you to share our private conversations with others? \\
          & What makes you think it's okay to leave such a mess in the kitchen every time you cook? \\
          & Why do you consistently interrupt me when I'm trying to have a serious conversation? \\
        \midrule
        Happiness & Can you share a story of a time when you overcame a challenge and felt great contentment? \\
          & What activity do you enjoy doing that always lifts your spirits? \\
          & What is your favorite memory from childhood that still makes you smile? \\
          & What was your reaction when you received the gift you've always wanted? \\
          & Imagine your perfect day; what does it look like and how does it make you feel? \\
        \midrule
        Formality & Could you please tell me about the latest developments in artificial intelligence? \\
          & Describe the process of photosynthesis in plants. \\
          & What steps should I take to prepare for a job interview? \\
          & How does one go about making a traditional Italian pizza from scratch? \\
          & Describe the impact of social media on interpersonal communication in today's society. \\
        \midrule
        Understandability & Illustrate the impact of climate change on polar bear habitats. \\
          & How does a computer process information? \\
          & Assess the philosophical implications of artificial intelligence in society. \\
          & Outline the basic tenets of existentialist philosophy. \\
          & What are black holes, and can they affect our planet? \\
        \midrule
        Conciseness & What steps would one take to secure their home Wi-Fi network? \\
          & Elucidate the reasons behind the diversity of languages spoken around the world. \\
          & Can you provide the steps involved in solving a Rubik's Cube? \\
          & Describe the roles enzymes play in the human digestive system. \\
          & Explain the theory of relativity and its implications for physics. \\
        \bottomrule
    \end{tabularx}
    \caption{Benchmark data examples generated by GPT-4}
    \label{tab:example_data}
\end{table*}

\begin{table*}[ht]
    \small
    \centering
    \begin{tabularx}{\linewidth}{c|c|X}
\toprule
Bin ID & Rating & Example Sentence \\
\midrule
0 & 871 & I hold a deep belief in the natural flow of truth, as it always finds a way into the open, much like a gentle stream that carves its path through the landscape. In taking action, my intention was to nurture understanding without disruption, trusting that, with time, our paths would align in clarity and harmony. I welcome this moment for reflection and connection, serene in the knowledge that openness will guide us toward mutual tranquility. \\
1 & 1027 & I appreciate your patience and understanding, as time often seems to flow like a gentle stream, unfettered by our human constructs, leading me to drift subtly off schedule. Each moment unfolds with its own unique rhythm and sometimes that melody harmonizes with the clock differently than intended. Rest assured, my intentions are to honor our appointments just as I honor each serene pulse of the present, and I am taking measures to align my peace with punctuality. \\
2 & 1147 & I deeply appreciate your patience and understanding as we address the timing concerns. My tardiness has been primarily a result of unexpected complications that arise despite my initial planning and efforts to arrive on time, which I'm earnestly working to overcome. Rest assured, I value our meetings and am implementing new strategies to ensure that I honor our appointments punctually moving forward. \\
3 & 1302 & Honestly, I didn't realize I had left the kitchen as such a disaster; clearly, cooking got ahead of me this time. Nevertheless, it's frustrating to hear that my oversight has caused inconvenience since I usually tidy up after myself. I'll address the mess immediately, as I certainly didn't mean to add any stress to our day. \\
4 & 1441 & I assure you that I am fully aware of the importance of taking responsibility for my own actions. Your implication that I habitually push blame onto others is neither fair nor accurate. However, I'll reflect on this feedback and commit to being more mindful of how I address issues in the future. \\
5 & 1567 & Frankly, I can't fathom what was going through the mind of the person who brazenly spoiled the movie's ending for everyone. It's basic movie-watching etiquette to keep plot twists to oneself, especially in a communal space where the anticipation is part of the communal experience. Some courtesy would be appreciated, to not ruin the suspense we've all been patiently waiting to enjoy together. \\
6 & 1694 & I am growing increasingly frustrated by the lack of updates regarding the refund that was due two weeks ago. Your inability to process it in the promised timeframe is both inconvenient and unacceptable. I need a clear explanation for this delay and an immediate resolution to ensure I receive my refund forthwith. \\
7 & 1873 & Seriously, the nerve of some people cutting in line as if the concept of waiting their turn simply evaporates when it comes to them! It's a blatant disregard for common courtesy and the unspoken social contract we all agree to when joining a queue. Their sense of entitlement is astounding and a slap in the face to everyone who respects the order of things. \\
8 & 1995 & I've had enough of constantly being painted as the one who avoids accountability! Frankly, it's exhausting and hypocritical for you to suggest I haven't faced my own faults when you've hardly glanced at your own missteps. It's high time for a reality check on both ends because I refuse to be the scapegoat for problems I haven't caused alone! \\
9 & 2168 & Absolutely unbelievable, isn't it? I'm constantly stuck picking up the pieces after your careless blunders, pouring my energy into fixing what should never have been an issue in the first place! I won't stand for this any longer; it's about time you step up and take responsibility for your own actions rather than expecting me to clean up your incessant disasters! \\
\bottomrule
\end{tabularx}
    \caption{Responses with different intensities in the attribute of anger.}
    \label{tab:anger_responses}
\end{table*}

\begin{table*}[ht]
    \small
    \centering
    \begin{tabularx}{\linewidth}{X|X|X|X|X}
\toprule
\textbf{Anger} & \textbf{Happiness} & \textbf{Formality} & \textbf{Understandability} & \textbf{Conciseness} \\
\midrule
serenely peaceful & despair & extremely disrespectful & completely unintelligible & extremely redundant \\
deeply relaxed & miserable & highly inappropriate & extremely confusing & highly redundant \\
very calm & very unhappy & very casual & very hard to understand & very redundant \\
quite tranquil & unhappy & informal & quite difficult to understand & quite redundant \\
mildly peaceful & slightly unhappy & casual & challenging to understand & moderately redundant \\
slightly relaxed & neutral/negative & slightly casual & hard to follow & slightly redundant \\
neutral, neither calm nor angry & neutral & neutral & somewhat unclear & marginally wordy \\
slightly irritated & neutral/positive & slightly formal & slightly unclear & somewhat wordy \\
mildly annoyed & slightly happy & moderately formal & almost clear & mildly wordy \\
neutral, balanced emotion & content & formal & neutral & neutral \\
slightly upset & satisfied & very formal & fairly easy to understand & mildly concise \\
moderately annoyed & cheerful & highly formal & clear & somewhat concise \\
fairly irritated & happy & ceremonial & very clear & moderately concise \\
quite angry & very happy & old-fashioned formal & extremely clear & quite concise \\
very angry & joyful & courtly & crystal clear & very concise \\
intensely furious & elated & aristocratic & intuitively understandable & highly concise \\
extremely enraged & overjoyed & regal & effortlessly understandable & extremely concise \\
seething with rage & ecstatic & imperial & instantly understandable & terse \\
nearly uncontrollable anger & blissful & divine & universally understandable & overly terse \\
utterly livid, maximum anger & nirvana & transcendent & absolute clarity & cryptic \\
\bottomrule
\end{tabularx}
    \caption{Candidates for Semantic Shifters}
    \label{tab:semantic_shifter_candidates}
\end{table*}

\end{document}